\documentclass[runningheads, envcountsame]{llncs}
\usepackage[pdftex]{graphicx}
\usepackage{epstopdf}
\usepackage{amsmath,amssymb,amsfonts,algorithm,enumerate,appendix,color}
\usepackage{microtype}
\usepackage{algpseudocode}
\usepackage{tabularx}
\usepackage{multirow}
\usepackage[colorlinks=true,linkcolor=blue]{hyperref}
\usepackage{bm}
\usepackage{url}
\usepackage[misc]{ifsym}

\title{Disentangled Sticky Hierarchical Dirichlet Process Hidden Markov Model}
\titlerunning{Disentangled Sticky Hierarchical Dirichlet Process Hidden Markov Model}
\toctitle{Disentangled Sticky Hierarchical Dirichlet Process Hidden Markov Model}
\author{Ding Zhou\inst{1}(\Letter) \and Yuanjun Gao\inst{2} \and Liam Paninski\inst{1}}
\authorrunning{D. Zhou, Y. Gao, and L. Paninski}
\tocauthor{Ding Zhou, Yuanjun Gao, and Liam Paninski}
\institute{Department of Statistics, Columbia University\\
\email{dz2336@columbia.edu, liam@stat.columbia.edu} \and Jane Street}

\begin{document}

\maketitle
\setcounter{footnote}{0}

\begin{abstract}
The Hierarchical Dirichlet Process Hidden Markov Model (HDP-HMM) has been used widely as a natural Bayesian nonparametric extension of the classical Hidden Markov Model for learning from sequential and time-series data. A sticky extension of the HDP-HMM has been proposed to strengthen the self-persistence probability in the HDP-HMM.  However, the sticky HDP-HMM entangles the strength of the self-persistence prior and transition prior together, limiting its expressiveness.
Here, we propose a more general model: the disentangled sticky HDP-HMM (DS-HDP-HMM). 
We develop novel Gibbs sampling algorithms for efficient inference in this model. We show that the disentangled sticky HDP-HMM outperforms the sticky HDP-HMM and HDP-HMM on both synthetic and real data, and apply the new approach to analyze neural data and segment behavioral video data.
\end{abstract}
\section{Introduction}

Hidden Markov models (HMMs) provide a powerful set of tools for modeling time series data. In the HMM we assume that the time series observations are modulated by underlying latent time-varying variables which take a discrete set of states. 
This model class is useful in its own right and can also be incorporated as a building block for more complicated models. It has been widely used in speech recognition \cite{fox2011sticky}, musical audio analysis \cite{pardo2005modeling,hoffman2008data}, acoustic-phonetic modeling \cite{lee2012nonparametric}, behavior segmentation \cite{fox2009nonparametric,saeedi2016segmented,batty2019behavenet}, sequential text modeling \cite{teh2006,heller2009infinite}, financial time series data analysis \cite{fox2009nonparametric,song2014modelling}, computational biology \cite{krogh1994hidden}, and many other fields.

Selecting the number of HMM states is an important question for practitioners. Classical model selection techniques can be used, but these methods can be computationally intensive and are sometimes unreliable in practice~\cite{celeux2008selecting,pohle2017selecting}. Also, for real datasets, it is often reasonable to assume that the number of latent states may be unbounded, violating classical assumptions needed to establish consistency results for model selection.  
Based on previous work in \cite{beal2002infinite}, \cite{teh2006} proposed the Hierarchical Dirichlet process HMM (HDP-HMM), a Bayesian nonparametric framework. 
In the HDP-HMM the transition matrix follows a hierarchical Dirichlet process (HDP) prior. \cite{fox2011sticky} noted that the  HDP-HMM tends to rapidly switch among redundant states, and proposed the sticky HDP-HMM (S-HDP-HMM), which strengthens the self-persistence probability. This modification often leads to significant improvements in modeling real data.

In the HMM, it is important to distinguish three features: 1, the 
similarity of the rows of the transition matrix; 2, the average self-persistence probability of the latent states (controlled by the mean of the diagonal of the transition matrix); and 3, the strength of the self-persistence prior across states (i.e., the inverse prior variance of the diagonal elements of the transition matrix).  In the HDP-HMM, there is only one parameter controlling feature 1. The sticky HDP-HMM adds one more parameter to control feature 2, but still entangles features 1 and 3 with only one parameter, thus limiting the expressiveness of the prior.

We show that we can add one additional hyperparameter 
to generalize the sticky HDP-HMM formulation, obtaining three degrees of freedom to model the three features discussed above. We call this new model the disentangled sticky HDP-HMM (DS-HDP-HMM).

The rest of the paper is organized as follows. In section \ref{sec:bg}, we provide a brief introduction to Bayesian HMM, HDP-HMM, and sticky HDP-HMM. In section \ref{sec:limitation}, we discuss the limitations of these models. In section \ref{sec:ds-hdp-hmm}, we introduce disentangled sticky HDP-HMM, and in section \ref{sec:gibbs} we develop efficient Gibbs sampling inference methods for this new model. Section \ref{sec:rlt} demonstrates the effectiveness of the disentangled sticky HDP-HMM on both synthetic and real data, including applications to analyzing neural data and segmenting behavior video. The notation table can be found in the supplementary material section~\ref{supp:notations}.

\section{Background on Bayesian HMM and HDP-HMM}
\label{sec:bg}

Our goal here is to fit an HMM to time series data.  On its face, this would seem to be a solved problem; after all, we can compute the HMM likelihood easily, and the basic  expectation-maximization algorithm for HMM fitting is textbook material \cite{rabiner1989tutorial}.
Nonetheless, a fully Bayesian solution to this problem has remained elusive. Specifically, we would like to be able to compute a posterior over all of the unknown HMM parameters (including the number of latent states).  Quantification of  posterior uncertainty is critical in many applications: for example, given short time series data, often we do not have enough data to sufficiently identify the HMM parameters.  Even for longer time series data, we might want to fit richer models as we collect more data.  Here ``richer models" correspond to more latent states, and since the number of parameters in the HMM grows  quadratically with the number of states, we may be left again with some irreducible uncertainty about the model parameters.





The HDP-HMM \cite{teh2006} provides a useful starting point for fully Bayesian HMM inference.
The basic idea here is to sample a global transition distribution prior from a Dirichlet process (described below), and then for each latent state we sample a transition distribution from this shared (random) global prior distribution.
To develop the details of this  HDP-HMM idea we first need to define some notation for the Dirichlet process (DP).  
Given a base distribution $H$ on a parameter space $\mathrm{\Theta}$ and a positive concentration parameter $\gamma$, a Dirichlet process $G\sim \mathrm{DP}(\gamma, H)$ (sometimes also denoted by $\mathrm{DP}(\gamma H)$) can be constructed by the following stick-breaking procedure \cite{sethuraman1994constructive}: let
\begin{equation}
    \beta \sim \mathrm{GEM}(\gamma),\  \theta_i \overset{iid}{\sim} H,\ i=1,2,\cdots,
\end{equation}
where $\beta \sim \mathrm{GEM}(\gamma)$ is a random probability mass function (p.m.f.) defined on a countably infinite set as follows:
\begin{equation}
\begin{aligned}
    v_i &\sim \mathrm{Beta}(1,\gamma),\ \beta_i = v_i\prod_{l=1}^{i-1}(1-v_l),\ i=1,2,\cdots.
\end{aligned}
\end{equation}
Then the discrete random measure $G = \sum_{i} \beta_i\delta_{\theta_i}$ is a sample from $\mathrm{DP}(\gamma H)$, where $\delta_{\theta_i}$ denotes the Dirac measure centered on $\theta_i$. 




The HDP-HMM \cite{teh2006} uses the DP to define a prior on the rows of the HMM transition matrix in a setting where the number of latent states is unbounded. 
The HDP-HMM is defined as
\begin{equation}
\begin{aligned}
    \mathrm{DP\ shared\ global\ prior:\quad}&\beta\sim \mathrm{GEM}(\gamma) \\
    &\theta_j\overset{iid}{\sim}H,\ j=1,2,\cdots \\
    \mathrm{Transition\ matrix\ prior:\quad}&\pi_j \overset{iid}{\sim} \mathrm{DP}(\alpha\beta),\ j=1,2,\cdots\\
    \mathrm{Latent\ states:\quad}&z_t\sim \pi_{z_{t-1}},\ t=1,\cdots,T\\
    \mathrm{Observations:\quad}&y_t\sim f(y|\theta_{z_t}),\ t=1,\cdots,T
\end{aligned}
\end{equation}
Here, $\beta$ and $\{\theta_j\}_{j=1}^{\infty}$ are defined as in the DP described above, and then each transition distribution $\pi_j$ for state $j$ is defined as a random sample of a second DP with base measure $\beta$ and concentration parameter $\alpha$. Here $\alpha$ controls how similar $\pi_j$ is to the global transition distribution $\beta$.  Finally, as usual, $z_t$ denotes the state of a Markov chain at time $t$, and the observation $y_t$ is independently distributed given the latent state $z_t$ and parameters $\{\theta_j\}_{j=1}^{\infty}$, with emission distribution $f(\cdot)$.

\begin{figure}[!ht]
    \centering
    \includegraphics[width=0.7\textwidth,clip=true]{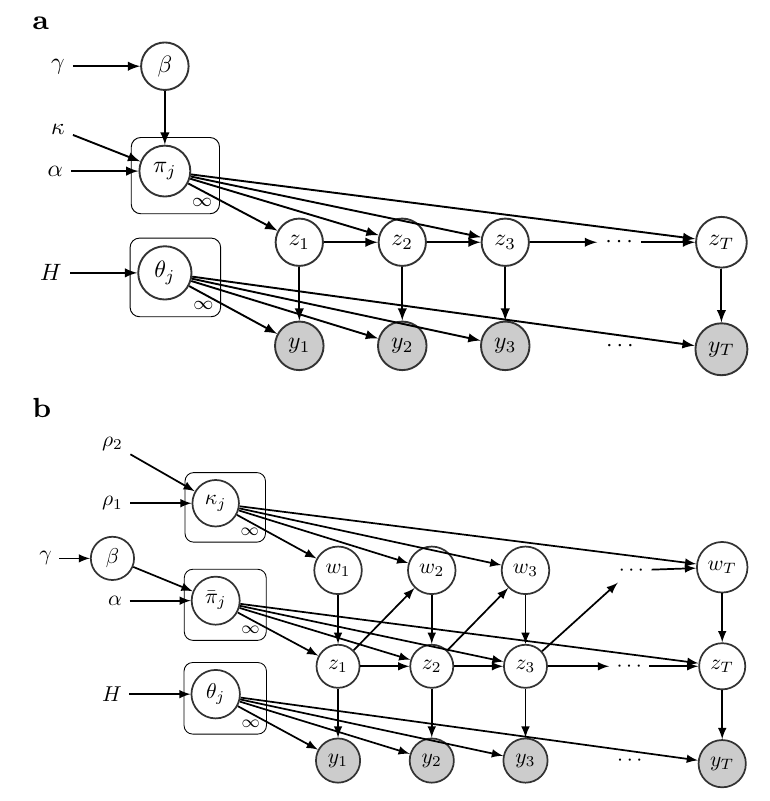}
    \caption{Graphical models for sticky HDP-HMM (a) and disentangled sticky HDP-HMM (b).}
    \label{fig:graph_shdphmm}
\end{figure}

The sticky HDP-HMM from \cite{fox2011sticky} modifies the transition matrix prior by adding a point mass distribution with stickiness parameter $\kappa$ to encourage self-persistence:
\begin{equation}\label{eq:efox}
     \mathrm{Transition\ matrix\ prior:\quad}\pi_j \sim \mathrm{DP}(\alpha\beta+\kappa\delta_j),\ j=1,2,\cdots,
\end{equation}
where $\delta_j$ denotes the Dirac measure centered on $j$. Figure~\ref{fig:graph_shdphmm}(a) provides the graphical model for the sticky HDP-HMM. 

\section{Limitations of the HDP-HMM and Sticky HDP-HMM}
\label{sec:limitation}

The HDP-HMM uses the concentration parameter $\alpha$ to control the following feature of the HMM:
\begin{itemize}
\item feature 1: the strength of the transition matrix prior, or the similarity of the rows of the transition matrix.
\end{itemize}
In other words, a large value of $\alpha$ here means that the transition probability for each state is close to the global transition distribution $\beta$.

A flexible model should have additional parameters to control two additional features:
\begin{itemize}
\item feature 2: the average self-persistence probability, or the mean of the diagonal of the transition matrix. 
\item feature 3: the strength of the self-persistence prior, or the similarity of the diagonal elements of the transition matrix.
\end{itemize}

The sticky HDP-HMM adds just one parameter $\kappa$ compared to the HDP-HMM. feature 2 is controlled by $\kappa / (\alpha + \kappa)$, while both feature 1 and feature 3 are controlled by $\alpha + \kappa$. Note that a strong prior (large $\alpha+\kappa$) means that both the self-persistence probability and the transition probability are quite similar across states, and it is impossible to separate the strength of these two elements using this parameterization.   These three features should occupy three degrees of freedom in total, but the formulation of sticky HDP-HMM is only able to traverse a two-dimensional sub-manifold of this three-dimensional space, 
limiting the expressiveness of the sticky HDP-HMM prior.

More concretely, consider the speaker diarization example studied in \cite{fox2011sticky}. The task is to distinguish the speakers in an audio recording of a conversation; the current speaker is the hidden state. Suppose that some speakers are likely to speak for a very long time while others are terse (implying a small feature 3, i.e. small $\alpha+\kappa$), but the identity of the next speaker is independent of the identity of the previous speaker (implying a big feature 1, i.e. big $\alpha+\kappa$). The sticky HDP-HMM would have trouble in this scenario. The opposite case is plausible as well: as an example, consider a group of people sitting in a circle and expressing their opinions in a clock-wise fashion (implying a small feature 1 since the distribution of the next speaker is highly dependent on the previous speaker); if each speaker talks for a very similar amount of time (implying a big feature 3), then the sticky HDP-HMM would have difficulty with this scenario as well.

\section{Disentangled Sticky HDP-HMM}
\label{sec:ds-hdp-hmm}

Now that we have diagnosed this lack of flexibility in the HDP-HMM and sticky HDP-HMM, we can construct a new more flexible model that
separates the strength of the self-persistence from the similarity of the transition probabilities.  Specifically, we modify the transition matrix prior as
\begin{equation}
\begin{aligned}
    \mathrm{Transition\ matrix\ prior:\quad}&\kappa_j\overset{iid}{\sim}\mathrm{beta}(\rho_1,\rho_2)
    \\
    &\bar{\pi}_j\overset{iid}{\sim}\mathrm{DP}(\alpha\beta)\\
    &\pi_j =\kappa_j\delta_j+ (1-\kappa_j)\bar{\pi}_j,\ j=1,2,\cdots,
\end{aligned}
\end{equation}
where the transition distribution $\pi_j$ is a mixture distribution. A sample from $\pi_j$ has the self-persistence probability of $\kappa_j$ to come from a point mass distribution at $j$, and has probability $1-\kappa_j$ to come from $\bar{\pi}_j$, a sample from the $\mathrm{DP}$ with base measure $\beta$. We call this new model the disentangled sticky HDP-HMM.

Here the $\mathrm{beta}(\rho_1,\rho_2)$ prior has the flexibility to control both the expectation of self-persistence (feature 2), and the variability of self-persistence (feature 3).
Meanwhile $\alpha$ is free to control the variability of the transition probability around the mean transition $\beta$ (feature 1). In short, we use 3 parameters $(\rho_1,\rho_2, \alpha)$ rather than the 2 parameters $(\kappa,\alpha)$ to separate the strength of the self-persistence and the transition priors.

When $\rho_1=0$ and $\rho_2>0$, all the $\kappa_j=0$, and the disentangled sticky HDP-HMM reduces to HDP-HMM. 
Importantly, the sticky HDP-HMM is also a special case of the disentangled sticky HDP-HMM, as shown in Theorem~\ref{thm}. See the supplementary material section~\ref{supp:proof} for a proof.

\begin{theorem}\label{thm}
The sticky HDP-HMM formulation in equation~\ref{eq:efox} is a special case of the disentangled sticky HDP-HMM by setting $(\rho_1,\rho_2)=(\kappa,\alpha)$.  
\end{theorem}


An equivalent formulation of $z_t\sim\pi_{z_{t-1}}$ in the disentangled sticky HDP-HMM is as follows:
\begin{equation}
\begin{aligned}
    \mathrm{Latent\ states:\quad}&w_t\sim \mathrm{Ber}(\kappa_{z_{t-1}})\\
    &z_t\sim w_t\delta_{z_{t-1}} + (1-w_t)\bar{\pi}_{z_{t-1}},\ t=1,\cdots,T,
\end{aligned}
\end{equation}
where we add binary auxiliary variables $w_{t}$, which decide whether the next step is self-persistent or switching. See Figure~\ref{fig:graph_shdphmm}(b) for the corresponding graphical model. We use this formulation to facilitate inference in section~\ref{sec:gibbs}.

\section{Gibbs Sampling Inference}
\label{sec:gibbs}

In this section, we 
introduce a new direct assignment Gibbs sampler (Algorithm~\ref{alg}) (similar to \cite{teh2006}) and a new weak-limit Gibbs sampler (Algorithm~\ref{alg:weak-limit}) (similar to \cite{fox2011sticky}) for inference in the disentangled sticky HDP-HMM. 
The direct assignment sampler generates samples from the true posterior of the disentangled sticky HDP-HMM when the Gibbs chains converge.  The weak-limit sampler uses finite approximation of the HDP-HMM to accelerate the mixing rate of the Gibbs chains and can be easily adapted to parallel computing. \cite{fox2009nonparametric} noted that the weak-limit sampler was useful for observation models with dynamics such as auto-regressive HMM (ARHMM) or switching linear dynamic system (SLDS). For detailed derivations of these two algorithms, see the supplementary material section~\ref{supp:gibbs}.

\subsection{Direct Assignment Sampler}
The direct assignment sampler for the HDP-HMM marginalizes transition distributions $\pi_j$ and parameters $\theta_j$ and sequentially samples $z_t$ given all the other states $z_{\setminus t}$, observations $\{y_{t}\}_{t=1}^T$, and the global transition distribution $\beta$.
The main difference between our direct assignment sampler and the corresponding  HDP-HMM sampler is that instead of only sampling $z_t$, we sample $\{z_t, w_t, w_{t+1}\}$ in blocks. We sample  $\alpha,\beta,\gamma$ only using $z_t$ that switch to other states by $\bar{\pi}_{z_{t-1}}$ ($w_t=0$), and sample $\{\kappa_j\}_{j=1}^{K+1},\rho_1,\rho_2$ only using $z_t$ that stick to state $z_{t-1}$ ($w_t=1$).

\begin{algorithm}[!ht]\label{alg}
\begin{algorithmic}[1]
    \item Sequentially sample $\{z_t, w_t, w_{t+1}\}$ for $t=1,\cdots,T$. 
    \item Sample $\{\kappa_j\}_{j=1}^{K+1}$. $K$ is defined as number of unique states in $\{z_t\}_{t=1}^T$. 
    \item Sample $\beta$. Same as HDP-HMM. 
    \item Optionally, sample hyperparameter $\alpha, \gamma, \rho_1, \rho_2$. 
\end{algorithmic}
\caption{Direct assignment sampler for disentangled sticky HDP-HMM}
\end{algorithm} 

For step 1, we sequentially compute the probability for each possible case of the posterior $p(z_t, w_t, w_{t+1} | z_{\backslash t}, w_{\backslash \{t, t+1\}}, \{y_t\}_{t=1}^T, \alpha, \beta, \{\kappa_j\}_{j=1}^{K+1})$, and sample $\{z_t, w_t, w_{t+1}\}$ from the corresponding multinomial distribution. If $z_t=K+1$, i.e. a new state appears, we will increment $K$, sample self-persistence probability $\kappa_{K+1}$ for a new state from the prior, and update $\beta$ using stick-breaking. For step 2, given $w_{t+1}$ whose corresponding $z_t$ is $j$, we can sample $\kappa_j$ using beta-binomial conjugacy.  For step 3, by introducing auxiliary variables $\{m_{jk}\}_{j,k=1}^K$, we  sample $\beta$ using Dirichlet-multinomial conjugacy. For step 4, we compute the empirical transition matrix $\{n_{jk}\}_{j,k=1}^K$, where $n_{jk}$ is the number of transitions from state $j$ to $k$ with $w_t=0$ in $\{z_t\}_{t=1}^T$, and introduce additional auxiliary variables. Then the posterior of $\alpha$ and $\gamma$ are gamma-conjugate, given the auxiliary variables. 
We approximate the posterior of $\rho_1,\rho_2$ by finite grids. 
The complexity for each step in Algorithm~\ref{alg} is $\mathcal{O}(TK)$, $\mathcal{O}(K)$, $\mathcal{O}(K)$, and $\mathcal{O}(K)$ respectively, so the total complexity per iteration is $\mathcal{O}(TK)$.

It is worth noting that instead of modeling $\{\kappa_j\}_{j=1}^{K+1}$ as samples from a beta distribution, it is natural to consider any distribution on the $[0,1]$ interval. The Gibbs algorithm here is easily adaptable to cases where we have extra prior information on the self-persistence probability.

\subsection{Weak-limit Sampler}
The weak-limit sampler for the sticky HDP-HMM constructs a finite approximation to the HDP prior based on the fact that 
\begin{equation}\label{eq:approx}
\begin{aligned}
    \beta|\gamma &\sim \mathrm{Dir}\left(\gamma/L,\cdots,\gamma/L\right)\\
    \pi_j|\alpha,\beta &\sim \mathrm{Dir}\left(\alpha\beta_1,\cdots,\alpha\beta_L\right),\ j=1,\cdots,L
\end{aligned}
\end{equation}
converges to the HDP prior when $L$ goes to infinity. Using this approximation, one can jointly sample latent variables $\{z_t\}_{t=1}^T$ with the HMM forward-backward procedure~\cite{rabiner1989tutorial}, which accelerates the mixing rate of the Gibbs sampler.

The main difference between our weak-limit Gibbs sampler and the corresponding sticky HDP-HMM sampler is that we now have two dimensional latent variables $\{z_t, w_t\}_{t=1}^T$ to sample. 

\begin{algorithm}[!ht]\label{alg:weak-limit}
\begin{algorithmic}[1]
    \item Jointly sample $\{z_t, w_t\}_{t=1}^T$.
    \item Sample $\{\kappa_j\}_{j=1}^L$.
    \item Sample $\{\beta_j\}_{j=1}^L,\{\bar{\pi}_j\}_{j=1}^L$. Same as HDP-HMM.
    \item Sample $\{\theta_j\}_{j=1}^L$.
    \item Optionally, sample hyperparameter $\alpha, \gamma, \rho_1, \rho_2$.
\end{algorithmic}
\caption{Weak-limit sampler for disentangled sticky HDP-HMM}
\end{algorithm}
For step 1, we apply the forward-backward procedure to jointly sample the two dimensional latent variables $\{z_t, w_t\}_{t=1}^T$.  Step 2 is the same as in Algorithm \ref{alg}. For step 3, we sample $\beta$ and $\bar{\pi}$ based on Dirichlet-multinomial conjugacy, given auxiliary variables $\{m_{jk}\}_{j,k=1}^L$, the empirical transition matrix $\{n_{jk}\}_{j,k=1}^L$, and the approximate prior in equation~\ref{eq:approx}. For step 4, we place a conjugate prior on $\theta_j$ and use conjugacy to sample from the posterior.
Step 5 is the same as in Algorithm \ref{alg}. 
The complexity for each step in Algorithm~\ref{alg:weak-limit} is $\mathcal{O}(TL^2)$, $\mathcal{O}(L)$, $\mathcal{O}(L)$, $\mathcal{O}(L)$, and $\mathcal{O}(L)$ respectively, with total complexity $\mathcal{O}(TL^2)$.

By jointly sampling the full latent sequence $\{z_t,w_t\}_{t=1}^T$, the weak-limit sampler greatly improves the mixing rate. The correlated observations in ARHMM and SLDS further slows the mixing rate of the direct assignment sampler, so jointly sampling is especially important for those models with dynamics~\cite{fox2009nonparametric}.

\section{Empirical Results}\label{sec:rlt}
In this section, we apply the disentangled sticky HDP-HMM to both simulated and real data.
We compared the performance of our disentangled sticky HDP-HMM with the two baseline models: sticky HDP-HMM and HDP-HMM. We evaluated the model performance according to two metrics: normalized Hamming distance (defined as the element-wise difference between the inferred states and the true underlying states) on training data, and the predictive negative log-likelihood on held-out test data.  To compute the Hamming distance, we used the Munkres algorithm \cite{munkres1957algorithms} to map the indices of the estimated state sequence to the set of indices that maximize the overlap with the true sequence. To compute the predictive negative log-likelihood, we used the set of parameters inferred every 10th Gibbs iteration after it converges, and ran the forward algorithm of \cite{rabiner1989tutorial}. 

For all the experiments, we adopted a full Bayesian approach, and used the following hyperpriors.  For all three models, we placed a $\mathrm{Gamma}(1, 0.01)$ prior on the concentration parameters $\alpha$ (and for the sticky HDP-HMM on $\alpha+\kappa$) to cover a wide range of $\alpha$ values.
We placed a $\mathrm{Gamma}(2,1)$ prior on $\gamma$ to avoid extremely small and large $\gamma$ samples, because small $\gamma$ will cause numerical instability when sampling $\beta$, while large $\gamma$ will generate too many states.
For our model and sticky HDP-HMM, we placed non-informative priors on the self-persistence parameters.  We placed a $\mathrm{Unif}([0,1])$ prior on the self-persistence proportion parameter $\phi=\frac{\rho_1}{\rho_1+\rho_2}$. For our model, we placed a $\mathrm{Unif}([0,2])$ on the self-persistence scale parameter $\eta = (\rho_1+\rho_2)^{-1/3},$ and cut $[0,1]\times[0,2]$ (the support of $\phi,\eta$) into $100\times 100$ grids (for simulated data) or $30\times 30$ (for real data) to numerically compute the posterior for $\phi,\eta$.

We used the direct assignment sampler to fit the simulated data, because the simulated data here has a relatively small sample size, and the direct assignment sampler generates samples from the true posterior of the model.  We used the weak-limit sampler to fit real data (with a larger sample size) in parallel.  We have also applied the direct assignment sampler to a short version of the hippocampal data;  the results obtained are qualitatively similar.

\subsection{Simulated Data}
In the simulation studies, we focus on two settings that serve to clearly illustrate the differences between the sticky versus disentangled sticky models.  In both settings, we experimented with both multinomial and Gaussian emissions. See the Gaussian emission results in the supplementary material section~\ref{supp:gaussian}.

\subsubsection{Different Self-persistence, Same Transition.} 
\begin{figure}[!ht]
    \centering
    \includegraphics[width = 1\textwidth,clip=true]{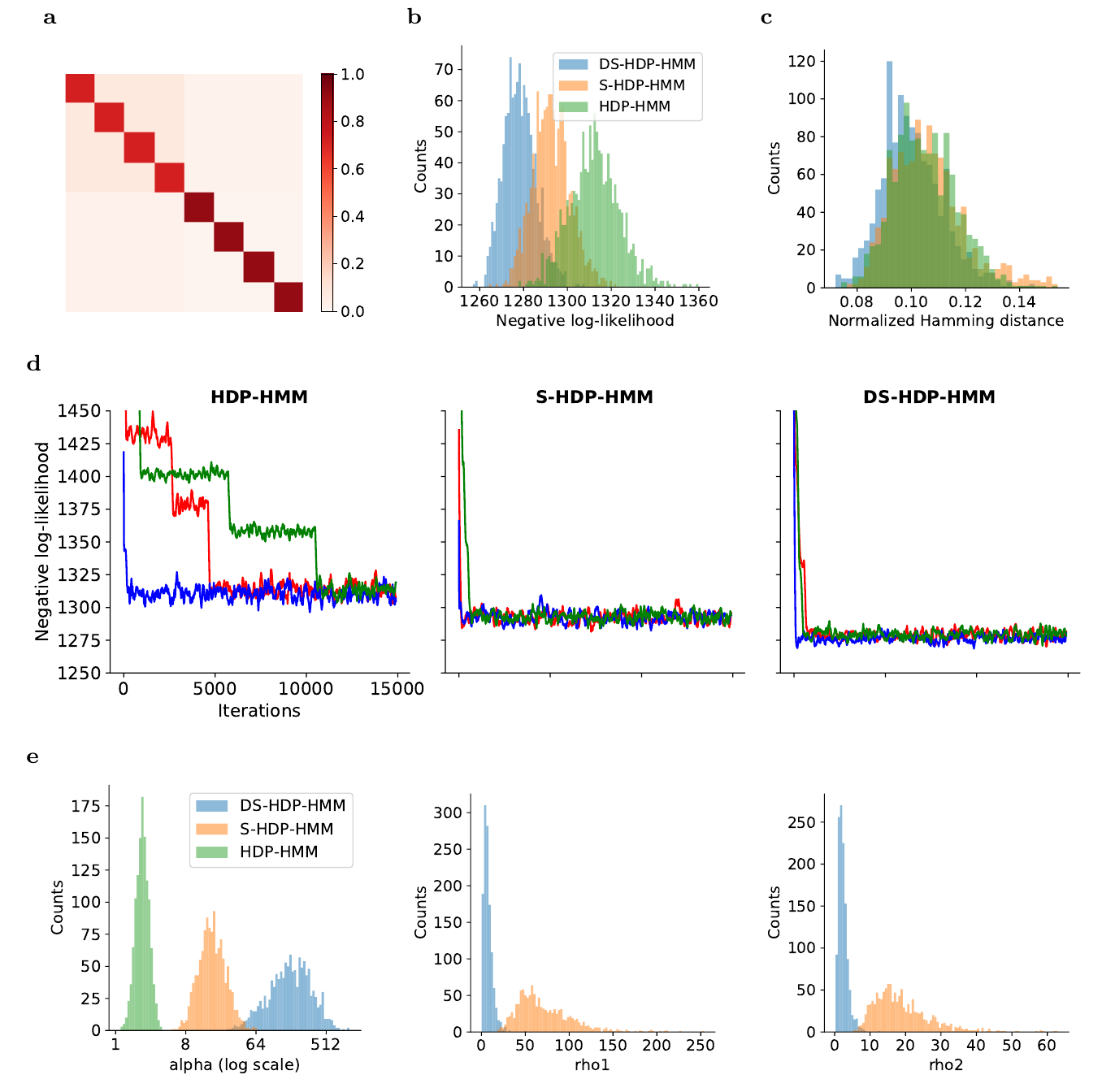}
    \caption{The DS-HDP-HMM provides good fits to the simulated data with different self-persistence ($\rho_1,\rho_2$ small), same transition ($\alpha$ large), and multinomial emission. (a) True transition matrix.  (b) Histogram of negative log-likelihood on test data.  (c) Histogram of normalized Hamming distance between the estimated states and true states on training data. 
    (d) Negative log-likelihood on test data over 15000 direct assignment Gibbs samples from 3 chains for each model. (e) Histogram of hyperparameters $\alpha, \rho_1,\rho_2$. Note that we plot $\kappa,\alpha$ for sticky HDP-HMM in the histogram of $\rho_1,\rho_2$, since $\kappa,\alpha$ would be equal to $\rho_1,\rho_2$ respectively if we treat sticky HDP-HMM as a special case of our model. Our model learns a big $\alpha$ and small $\rho_1,\rho_2$, which is consistent with the data. 
    }
    \label{fig:multinomial_same_trans}
\end{figure}

We simulated data using a transition matrix (Figure~\ref{fig:multinomial_same_trans}(a)) with different $\kappa_j$ and the same $\bar{\pi}_j$ across states,
which corresponds to a large transition concentration parameter $\alpha$ (big feature 1), and small self-persistence parameters $\rho_1,\rho_2$ (small feature 3).
We assigned the multinomial observation to be equal to the latent state with probability $0.9$ and to other states with equal chance with probability $0.1$.

We placed a symmetric $\mathrm{Dir}(1,\cdots,1)$ prior on the multinomial parameters. We ran $3$ MCMC chains, each with 15000 iterations, with 11000 iterations as burn-in (Figure~\ref{fig:multinomial_same_trans}(d)).

Results of three models fit on multinomial emission are shown in Figure~\ref{fig:multinomial_same_trans}. As shown in Figure~\ref{fig:multinomial_same_trans}(e), our model learns a big $\alpha$ and small $\rho_1,\rho_2$, which is consistent with the data, while sticky HDP-HMM model entangles these two parameters and learns something in the middle.  Better recovery of the hyperparameters leads to better fits on the data: our model outperforms the two baseline models in terms of negative log-likelihood and Hamming distance (Figure~\ref{fig:multinomial_same_trans}(b)(c)).  The advantages of our model are even more clear under the Gaussian emission (see figure in the supplementary material section~\ref{supp:gaussian}). The three models generally learn similar $\gamma$, which means that they infer similar numbers of states. 

For each emission model, we compared these three models on 10 datasets generated from the same model using different random seeds.  The conclusions are consistent across the 10 different datasets.

\subsubsection{Same Self-persistence, Different Transition.}
\begin{figure}[!ht]
    \centering
    \includegraphics[width = 1\textwidth,clip=true]{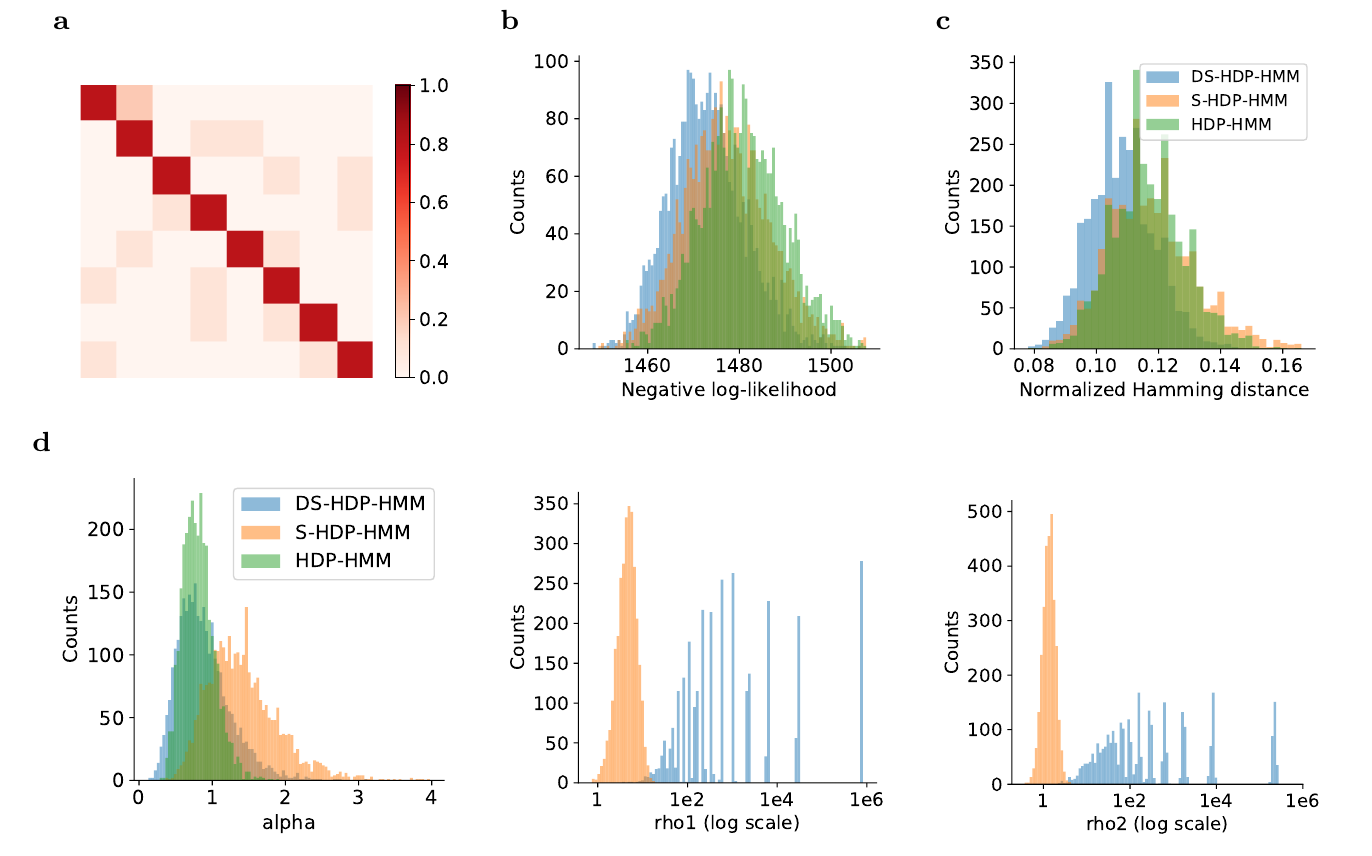}
    \caption{Results for the simulated data with same self-persistence ($\rho_1,\rho_2$ large), different transition ($\alpha$ small) and multinomial emission. We ran 3 chains, each with 30000 iterations, with 20000 iterations as burn-in for this data. Conventions as in  Figure~\ref{fig:multinomial_same_trans}. Our model learns a small $\alpha$ and big $\rho_1,\rho_2$, which is consistent with the data. Note that the ``spikiness'' in the histogram of $\rho_1,\rho_2$ comes from the approximation of the posterior of $\rho_1,\rho_2$ using finite grids. An alternative way to avoid the discretization would be to use 
    Metropolis-Hasting sampling~\cite{hastings1970monte}.}
    \label{fig:multinomial_diff_trans}
\end{figure}

We simulated data using a transition matrix (Figure~\ref{fig:multinomial_diff_trans}(a)) with the same $\kappa_j$ and different $\bar{\pi}_j$ across states,
which corresponds to small $\alpha$ (small feature 1), and large $\rho_1,\rho_2$ (large feature 3). 
We assigned the multinomial observation to be equal to the latent state with probability 0.8 and to other states with equal chance with probability 0.2.

Again, consistently across 10 replications, our model learns hyperparameters consistent with the data (Figure~\ref{fig:multinomial_diff_trans}(d)) and outperforms the two baseline models (Figure~\ref{fig:multinomial_diff_trans}(b)(c)), though the advantages are not as big as in the previous scenario. This is likely because in the previous example, our model can learn the hyperparameters based on the similarity among rows of the transition matrix, while in this scenario, it can learn hyperparameters mostly from the similarity among diagonal elements of the transition matrix, which contain much less information.

\subsection{Inferring Rat Hippocampal Population Codes}
\begin{figure}[!ht]
    \centering
    \includegraphics[width=1\textwidth,clip=true]{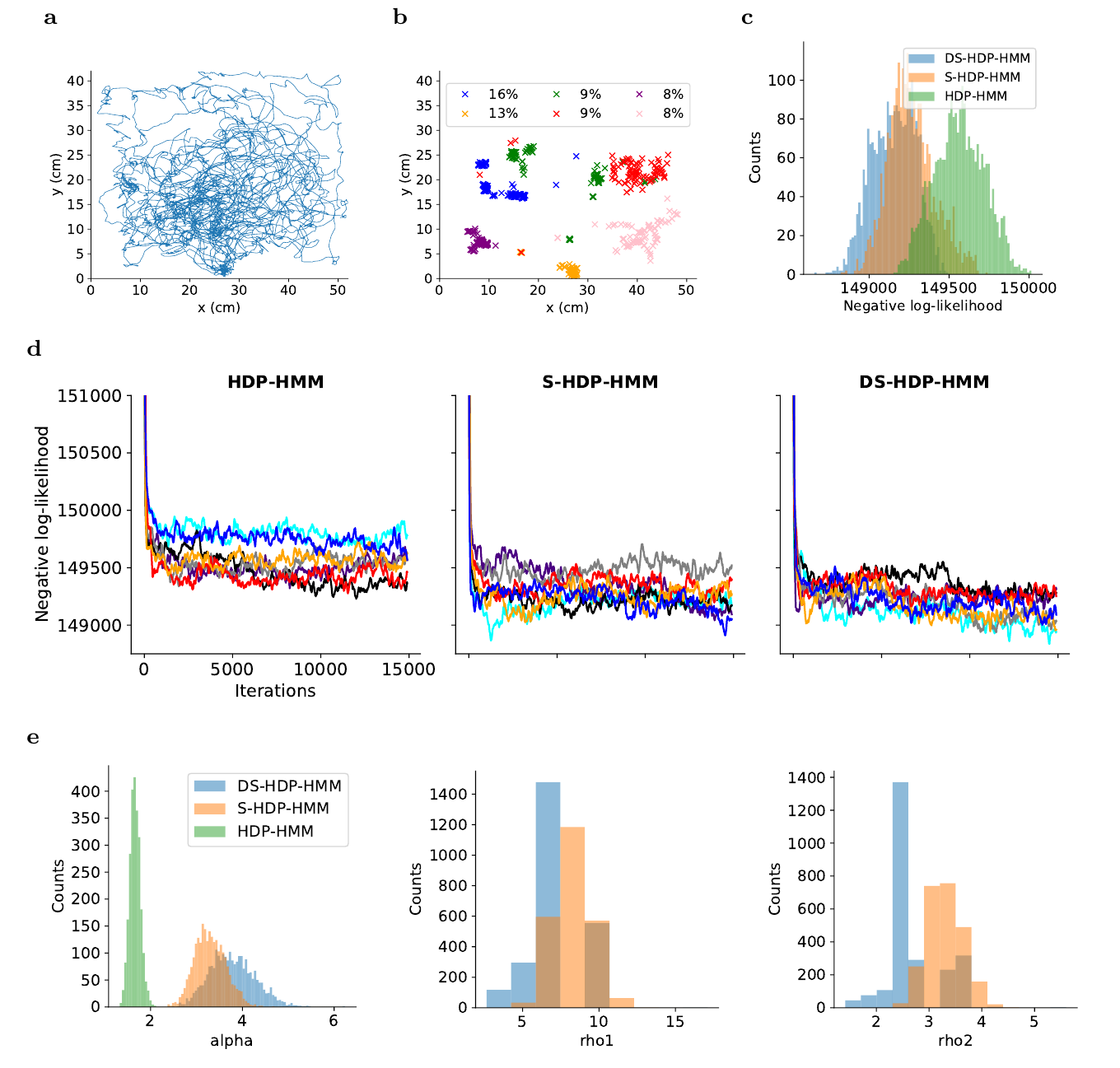}
    \caption{Results for rat hippocampal data. (a) The rat's moving trajectory in training and test data. (b) The spatial locations of the rat corresponding to $6$ most frequent states inferred by DS-HDP-HMM. The figure legend shows the percentage of each of the $6$ states. (c)(d)(e) Conventions as Figure~\ref{fig:multinomial_same_trans}, though with more traces and using the weak-limit sampler in (d).  Our model infers a bigger $\alpha$ than $\rho_1,\rho_2$ (big feature 1, small feature 3). }
    \label{fig:real_2d}
\end{figure}

Next we applied our model to a public electrophysiological hippocampus dataset \cite{pastalkova2015simultaneous,pastalkova2008internally}\footnote{\url{https://buzsakilab.nyumc.org/datasets/PastalkovaE/i01/i01_maze15_MS.001/}}.  In the experiment, a rat freely explored in an open square environment ($\sim50$cm$\times40$cm), while neural activity in the hippocampal CA1 area was recorded using silicon probes. See Figure~\ref{fig:real_2d}(a) for an example trace illustrating the position  of the rat over the course of the experiment.

We selected the $100$ most active putative pyramidal neurons and binned the ensemble spike activities with a frame rate of 10Hz. The dataset consists of $\sim 36$k frames. We used the first $8$k frames, cut it into blocks of $500$ frames, and randomly took 8 blocks as training data ($4$k frames), and the remaining 8 blocks as test data.  The spike count at time $t$ for cell $c$ is modeled as Poisson with rate $\lambda_{z_{t},c}$, i.e. $y_{t,c} \sim\mathrm{Poisson}(\lambda_{z_{t},c})$. As in \cite{linderman2016bayesian}, we used a conjugate $\mathrm{gamma}(a_c, b_c)$ prior for the firing rate $\lambda_{j,c},\ j=1,2,\cdots, \ c=1,\cdots,100$. We fixed the shape parameter $a_c=1$ across cells, and placed a $\mathrm{gamma}(1,1)$ prior for the scale parameter $b_c$.


We set $L=200$ and ran 7 MCMC chains, each with 15000 iterations, with 11000 iterations as burn-in (Figure~\ref{fig:real_2d}(d)).  Our model achieves the smallest negative log-likelihood on test data (Figure~\ref{fig:real_2d}(c)). A two-sample t-test for the mean difference between  negative log-likelihood of DS-HDP-HMM and S-HDP-HMM for $7$ chains is significant ($p$-value $<0.05$). The inferred states are correlated with the spatial locations of the rat (Figure~\ref{fig:real_2d}(b)).  Compared to sticky HDP-HMM, our model infers a bigger $\alpha$ and smaller $\rho_1, \rho_2$ (Figure~\ref{fig:real_2d}(e)). Bigger $\alpha$ implies smaller variability of the switching transition, consistent with the rat quickly going to other locations since it has fast running speed.  Smaller $\rho_1, \rho_2$ imply bigger variability of the self-persistence probability, consistent with the rat spending different durations at different locations.
Note that the disentangled sticky HDP-HMM and sticky HDP-HMM perform similarly on the rat hippocampal data; bigger differences are seen in the mouse behavior video data in the next section.

\subsection{Segmenting Mouse Behavior Video}

We also applied our model to a public mouse behavior dataset~\cite{churchland2019single,musall2019single}. In this experiment, a head-fixed mouse performed a visual decision task while neural activity across dorsal cortex was optically recorded using widefield calcium imaging. We only used the behavior video data (128x128 pixels each grayscale video frame), which was recorded using two cameras (one side view and one bottom view). The behavior video is high dimensional, so we directly adopted the dimension reduction result in~\cite{batty2019behavenet}, which output $9$-dimensional continuous variables estimated using a convolutional autoencoder.  

The dataset consists of 1126 trials across two sessions, with 189 frames each trial (30 Hz frame rate).  We randomly chose $100$ trials as training data ($\sim 20$k frames), and $30$ trials as test data ($\sim 6$k frames).  We assumed the observation follows an ARHMM, i.e. $y_t \sim \mathcal{N}\left(A_{z_t}y_{t-1}, \Sigma_{z_t}\right)$.  As in \cite{fox2009nonparametric}, we standardized the observations, and assumed $\{A_j,\Sigma_j\}$ follows a conjugate matrix-normal inverse-Wishart (MNIW) prior.  The details of this prior are in the supplementary material section~\ref{supp:arhmm}. We initialized the states $\{z_{t}\}_{t=1}^T$ of the three Bayesian nonparametric models using the state assignments result of a $32$-states parametric ARHMM in \cite{batty2019behavenet}. 

\begin{figure}[!ht]
    \centering
    \includegraphics[width=1\textwidth,clip=true]{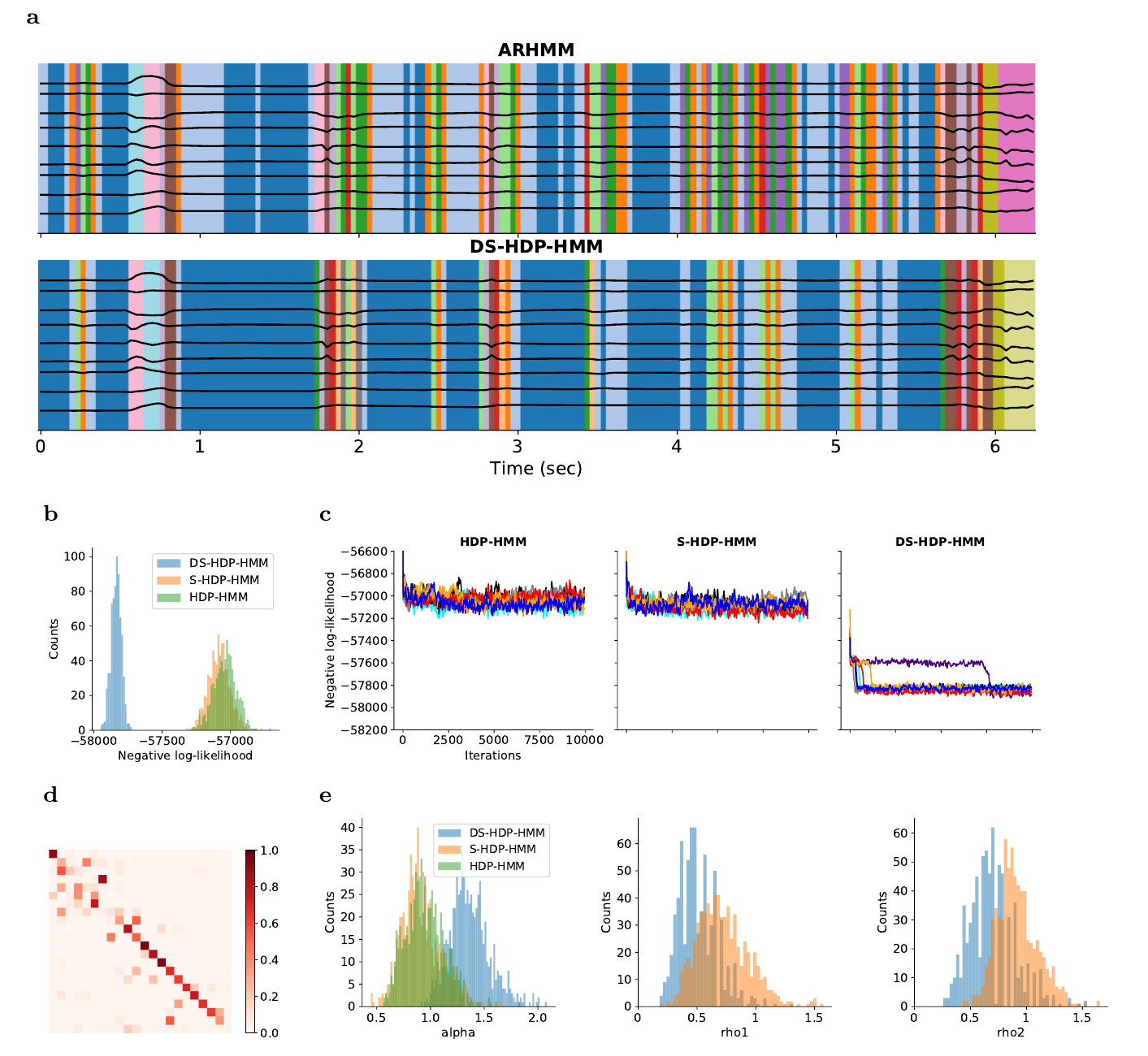}
    \caption{Results for mouse behavior data. (a) Observations (black) are shown on a sample trial over time, with background colors indicating the discrete state inferred for that time step using the ARHMM (number of states = 16) and our model (inferred number of states = 22) (colors are chosen to maximize the overlap between these two results). (b)(c)(e) Conventions as in the corresponding plots in  Figure~\ref{fig:real_2d}.  Note the significant loglikelihood gap between the DS-HDP-HMM and the baseline models here.  Our model infers a bigger $\alpha$ than $\rho_1,\rho_2$ (big feature 1, small feature 3). (d)  Transition matrix estimated by DS-HDP-HMM. Note that the diagonal elements are strongly variable across states (small feature 3).}
    \label{fig:real_ar}
\end{figure}

We set $L=40$ and ran 7 MCMC chains, each with 10000 iterations, with 9000 iterations as burn-in (Figure~\ref{fig:real_ar}(c)).
From Figure~\ref{fig:real_ar}(a), we can see that our model has less rapid switches among states than the parametric ARHMM even if we have more states. The diagonal elements are strongly variable across states (small feature 3) (Figure~\ref{fig:real_ar}(d)). For this dataset, $\alpha,\rho_1$, and $\rho_2$ are all small, but they still have different values. Our model infers a bigger $\alpha$ than $\rho_1,\rho_2$ (big feature 1, small feature 3), as shown in Figure~\ref{fig:real_ar}(e). Again, our model achieves much smaller negative predictive log-likelihood on test data than the other two Bayesian nonparametric models (Figure~\ref{fig:real_ar}(b)), due to disentanglement of hyperparameters, improving the expressiveness of the HDP prior.

\section{Discussion and Conclusion}
\label{conclusion}
In this paper, we propose an extension of the sticky HDP-HMM, to decouple the strength of the self-persistence prior (or the similarity of the diagonal elements of the transition matrix) and transition prior (or the similarity of the rows of the transition matrix). 
We develop two novel Gibbs samplers for performing efficient inference.
We also show in simulated and real data that our extension outperforms existing approaches.

The work \cite{johnson2013hdphsmm} proposed a Bayesian nonparametric prior to Hidden Semi-Markov Model \cite{murphy2002hidden} which also extends the sticky HDP-HMM. The idea is to break the Markovian assumption and instead model the distribution of duration of self-persistence for each state explicitly. The flexibility of the choice of duration distribution provides a rich set of possible priors, though the choice of duration distribution can be challenging for practitioners. Our formulation can be thought as a special case of their model, but with a few important advantages.
First, we clarify that the limitations  of the sticky HDP-HMM might not necessarily come from the Markovian assumption but could instead be due to the entanglement of the prior. Second, our extension adds a single hyperparameter,
making it easy for practitioners to use. Third, our formulation still maintains the Markovian structure, which is useful in many applications.
For example, there are applications of HDP-HMM prior to Markov decision processes \cite{doshi2009infinite,doshi2010nonparametric,doshi2013bayesian}, in which the Markovian structure is critical. Also, by harnessing the Markovian structure, our model enjoys scalable $O(TK^2)$ message passing algorithms, while the message passing algorithm of HSMM has complexity of $O(T^2K+TK^2)$ in general. 


In the future we hope to explore alternative computational approaches (specifically, stochastic variational \cite{hughes2015reliable,zhang2016stochastic} and amortized inference \cite{pakman2018discrete} methods), which may further improve the scalability of the model introduced here.  Another important direction is to adapt the mixture of finite mixtures (MFM) model \cite{miller2018mixture} to the HMM setting, as a replacement for the HDP prior; as in the mixture modeling setting discussed in \cite{miller2018mixture}, we expect that the MFM prior may lead to better estimates of the number of latent states. 

Open source code is available at \url{https://github.com/zhd96/ds-hdp-hmm}.

\section*{Acknowledgements}
We thank John Paisley, Xue-Xin Wei, Kenneth Kay, Matthew R Whiteway, Pengcheng Zhou and Ari Pakman for helpful discussions.  We thank the authors of \cite{batty2019behavenet} for generously sharing their data. We acknowledge computing resources from Columbia University's Shared Research Computing Facility project, which is supported by NIH Research Facility Improvement Grant 1G20RR030893-01, and associated funds from the New York State Empire State Development, Division of Science Technology and Innovation (NYSTAR) Contract C090171, both awarded April 15, 2010.

\bibliographystyle{splncs04}
\bibliography{dshdphmm}

\begin{thebibliography}{10}
\providecommand{\url}[1]{\texttt{#1}}
\providecommand{\urlprefix}{URL }
\providecommand{\doi}[1]{https://doi.org/#1}

\bibitem{batty2019behavenet}
Batty, E., Whiteway, M., Saxena, S., Biderman, D., Abe, T., Musall, S., Gillis,
  W., Markowitz, J., Churchland, A., Cunningham, J.P., et~al.: Behavenet:
  nonlinear embedding and bayesian neural decoding of behavioral videos. In:
  Advances in Neural Information Processing Systems. pp. 15680--15691 (2019)

\bibitem{beal2002infinite}
Beal, M.J., Ghahramani, Z., Rasmussen, C.E.: The infinite hidden markov model.
  In: Advances in neural information processing systems. pp. 577--584 (2002)

\bibitem{celeux2008selecting}
Celeux, G., Durand, J.B.: Selecting hidden markov model state number with
  cross-validated likelihood. Computational Statistics  \textbf{23}(4),
  541--564 (2008)

\bibitem{churchland2019single}
Churchland, A., Musall, S., Kaufman, M.T., Juavinett, A., Gluf, S.:
  Single-trial neural dynamics are dominated by richly varied movements:
  dataset  (2019). \doi{10.14224/1.38599}

\bibitem{doshi2009infinite}
Doshi-Velez, F.: The infinite partially observable markov decision process. In:
  Advances in neural information processing systems. pp. 477--485 (2009)

\bibitem{doshi2013bayesian}
Doshi-Velez, F., Pfau, D., Wood, F., Roy, N.: Bayesian nonparametric methods
  for partially-observable reinforcement learning. IEEE transactions on pattern
  analysis and machine intelligence  \textbf{37}(2),  394--407 (2013)

\bibitem{doshi2010nonparametric}
Doshi-Velez, F., Wingate, D., Roy, N., Tenenbaum, J.B.: Nonparametric bayesian
  policy priors for reinforcement learning. In: Advances in Neural Information
  Processing Systems. pp. 532--540 (2010)

\bibitem{escobar1995bayesian}
Escobar, M.D., West, M.: Bayesian density estimation and inference using
  mixtures. Journal of the american statistical association  \textbf{90}(430),
  577--588 (1995)

\bibitem{fox2009nonparametric}
Fox, E., Sudderth, E.B., Jordan, M.I., Willsky, A.S.: Nonparametric bayesian
  learning of switching linear dynamical systems. In: Advances in neural
  information processing systems. pp. 457--464 (2009)

\bibitem{fox2011sticky}
Fox, E.B., Sudderth, E.B., Jordan, M.I., Willsky, A.S.: A sticky hdp-hmm with
  application to speaker diarization. The Annals of Applied Statistics pp.
  1020--1056 (2011)

\bibitem{gelman2013bayesian}
Gelman, A., Carlin, J.B., Stern, H.S., Dunson, D.B., Vehtari, A., Rubin, D.B.:
  Bayesian data analysis. Chapman and Hall/CRC (2013)

\bibitem{hastings1970monte}
Hastings, W.K.: Monte carlo sampling methods using markov chains and their
  applications. Biometrika  \textbf{57}(1),  97--109 (1970)

\bibitem{heller2009infinite}
Heller, K., Teh, Y.W., Gorur, D.: Infinite hierarchical hidden markov models.
  In: Artificial Intelligence and Statistics. pp. 224--231 (2009)

\bibitem{hoffman2008data}
Hoffman, M.D., Cook, P.R., Blei, D.M.: Data-driven recomposition using the
  hierarchical dirichlet process hidden markov model. In: ICMC. Citeseer (2008)

\bibitem{hughes2015reliable}
Hughes, M., Kim, D.I., Sudderth, E.: Reliable and scalable variational
  inference for the hierarchical dirichlet process. In: Artificial Intelligence
  and Statistics. pp. 370--378 (2015)

\bibitem{johnson2013hdphsmm}
Johnson, M.J., Willsky, A.S.: {B}ayesian nonparametric hidden semi-{M}arkov
  models. Journal of Machine Learning Research  \textbf{14},  673--701
  (February 2013)

\bibitem{krogh1994hidden}
Krogh, A., Brown, M., Mian, I.S., Sjolander, K., Haussler, D.: Hidden markov
  models in computational biology. applications to protein modeling. Journal of
  molecular biology  \textbf{235}(5),  1501--1531 (1994)

\bibitem{lee2012nonparametric}
Lee, C.y., Glass, J.: A nonparametric bayesian approach to acoustic model
  discovery. In: Proceedings of the 50th Annual Meeting of the Association for
  Computational Linguistics: Long Papers-Volume 1. pp. 40--49. Association for
  Computational Linguistics (2012)

\bibitem{linderman2016bayesian}
Linderman, S.W., Johnson, M.J., Wilson, M.A., Chen, Z.: A bayesian
  nonparametric approach for uncovering rat hippocampal population codes during
  spatial navigation. Journal of neuroscience methods  \textbf{263},  36--47
  (2016)

\bibitem{miller2018mixture}
Miller, J.W., Harrison, M.T.: Mixture models with a prior on the number of
  components. Journal of the American Statistical Association
  \textbf{113}(521),  340--356 (2018)

\bibitem{munkres1957algorithms}
Munkres, J.: Algorithms for the assignment and transportation problems. Journal
  of the society for industrial and applied mathematics  \textbf{5}(1),  32--38
  (1957)

\bibitem{murphy2002hidden}
Murphy, K.P.: Hidden semi-markov models (hsmms). unpublished notes  \textbf{2}
  (2002)

\bibitem{murphy2007conjugate}
Murphy, K.P.: Conjugate bayesian analysis of the gaussian distribution. def
  \textbf{1}(2$\sigma$2), ~16 (2007)

\bibitem{musall2019single}
Musall, S., Kaufman, M.T., Juavinett, A.L., Gluf, S., Churchland, A.K.:
  Single-trial neural dynamics are dominated by richly varied movements. Nature
  neuroscience  \textbf{22}(10),  1677--1686 (2019)

\bibitem{pakman2018discrete}
Pakman, A., Wang, Y., Mitelut, C., Lee, J., Paninski, L.: Discrete neural
  processes. arXiv preprint arXiv:1901.00409  (2018)

\bibitem{pardo2005modeling}
Pardo, B., Birmingham, W.: Modeling form for on-line following of musical
  performances. In: Proceedings of the National Conference on Artificial
  Intelligence. vol.~20, p.~1018. Menlo Park, CA; Cambridge, MA; London; AAAI
  Press; MIT Press; 1999 (2005)

\bibitem{pastalkova2015simultaneous}
Pastalkova, E., Wang, Y., Mizuseki, K., Buzs{\'a}ki, G.: Simultaneous
  extracellular recordings from left and right hippocampal areas ca1 and right
  entorhinal cortex from a rat performing a left/right alternation task and
  other behaviors. CRCNS.org  (2015). \doi{10.6080/K0KS6PHF}

\bibitem{pastalkova2008internally}
Pastalkova, E., Itskov, V., Amarasingham, A., Buzs{\'a}ki, G.: Internally
  generated cell assembly sequences in the rat hippocampus. Science
  \textbf{321}(5894),  1322--1327 (2008)

\bibitem{pohle2017selecting}
Pohle, J., Langrock, R., van Beest, F., Schmidt, N.M.: Selecting the number of
  states in hidden markov models-pitfalls, practical challenges and pragmatic
  solutions. arXiv preprint arXiv:1701.08673  (2017)

\bibitem{rabiner1989tutorial}
Rabiner, L.R.: A tutorial on hidden markov models and selected applications in
  speech recognition. Proceedings of the IEEE  \textbf{77}(2),  257--286 (1989)

\bibitem{saeedi2016segmented}
Saeedi, A., Hoffman, M., Johnson, M., Adams, R.: The segmented ihmm: A simple,
  efficient hierarchical infinite hmm. In: International Conference on Machine
  Learning. pp. 2682--2691 (2016)

\bibitem{sethuraman1994constructive}
Sethuraman, J.: A constructive definition of dirichlet priors. Statistica
  sinica pp. 639--650 (1994)

\bibitem{song2014modelling}
Song, Y.: Modelling regime switching and structural breaks with an infinite
  hidden markov model. Journal of Applied Econometrics  \textbf{29}(5),
  825--842 (2014)

\bibitem{teh2006}
Teh, Y.W., Jordan, M.I., Beal, M.J., Blei, D.M.: Hierarchical dirichlet
  processes. Journal of the American Statistical Association
  \textbf{101}(476),  1566--1581 (2006). \doi{10.1198/016214506000000302}

\bibitem{tu2014dirichlet}
Tu, S.: The dirichlet-multinomial and dirichlet-categorical models for bayesian
  inference. Computer Science Division, UC Berkeley  (2014)

\bibitem{zhang2016stochastic}
Zhang, A., Gultekin, S., Paisley, J.: Stochastic variational inference for the
  hdp-hmm. In: Artificial Intelligence and Statistics. pp. 800--808 (2016)

\end{thebibliography}

\section*{Supplementary Material}
\appendix
\section{Notation}\label{supp:notations}
Table~\ref{tab:notations} summarizes the notation used in the main paper.
\begin{table}
\caption{Notation}\label{tab:notations}
\centering
\begin{tabular}{ll}
 \hline
 \multicolumn{2}{l}{General notation}\\
 \hline
 $x_{1:T}$ & the sequence $\{x_1,\cdots,x_T\}$\\
 $x_{\backslash t}$ & the sequence $\{x_1,\cdots,x_{t-1}, x_{t+1},\cdots, x_T\}$\\
 $x_{\cdot j}$ & $\sum_{i}x_{ij}$\\
 $x_{i\cdot}$ & $\sum_{j}x_{ij}$\\
 $\gamma$ & concentration parameter in DP prior \\ 
 $H$ & prior distribution of parameters \\
 $\mathrm{DP}(\gamma H)$ & Dirichlet process distribution with concentration parameter $\gamma$ and \\
 \ & base measure $H$ \\
 $\mathrm{GEM}(\gamma)$ &  stick-breaking distribution with parameter $\gamma$\\
 $\beta$ & global transition distribution \\
 $\alpha$ & concentration parameter in transition matrix prior\\
 $\theta_j$ & parameter of distribution of observations associated with state $j$\\
 $\pi_j$ & transition distribution for state $j$\\
 $z_t$ & latent state at time $t$ \\
 $y_t$ & observation at time $t$ \\
 \hline
 \multicolumn{2}{l}{Sticky HDP-HMM}\\
 \hline
 $\kappa$ & scalar stickiness parameter \\
 \hline
 \multicolumn{2}{l}{Disentangled sticky HDP-HMM}\\
 \hline
 $\kappa_j$ & self-persistence probability for state $j$\\
 $\rho_1,\rho_2$ & parameters in prior distribution of self-persistence probability\\
 $\bar{\pi}_j$ & transition distribution if not self-persistent for state $j$\\
 $w_t$ & binary variable indicating next step is self-persistent or switching\\
 $n_{jk}$ & number of transitions from state $j$ to $k$ with $w_t=0$ in $\{z_t\}_{t=1}^T$\\
 $n_{jk}^{-t}$ & number of transitions from state $j$ to $k$ with $w_t=0$ in $\{z_t\}_{t=1}^T$,\\
 & not counting $z_{t-1}$ to $z_t$ or $z_t$ to $z_{t+1}$\\
 $m_{jk}$ & number of tables in restaurant $j$ serving dish $k$ \\
 \hline
\end{tabular}
\end{table}
\section{Proof of Theorem~\ref{thm}}\label{supp:proof}
\begin{proof}
The stick breaking formulation of Dirichlet process gives us the following sampling strategy of $\mathrm{DP}(\beta')$ with $\beta'=(\kappa,\alpha\beta_1,\alpha\beta_2,\cdots)$:

\begin{gather*}
\eta'_0\sim\mathrm{beta}(\kappa,\alpha),\ \eta'_i\sim\mathrm{beta}(\alpha\beta_i,\ \alpha(1-\sum_{l\leq i}\beta_l)),\ i\geq1\\    
\eta_0 = \eta'_0,\ \eta_1 = \eta'_1,\ \eta_i=\eta'_i\prod_{1\leq l\leq i-1}(1-\eta'_l)
\end{gather*}

Now we have $\left(\eta_0,(1-\eta_0)\eta_1,\cdots,(1-\eta_0)\eta_i,\cdots,\right)\sim\mathrm{DP}(\beta')$.  As a result, $\left(\eta_0+(1-\eta_0)\eta_1,(1-\eta_0)\eta_2,\cdots\right)\sim\mathrm{DP}(\alpha\beta+\kappa\delta_1)$.  The exchangeability extends the results to general $\delta_j$ for $j\geq 1$.  On the other hand, we can interpret this construction as follows. Let $\eta = \left(\eta_1,\cdots,\eta_k,\cdots\right)$, then $\eta\sim\mathrm{DP}(\alpha\beta)$.  Note that $\eta_0$ is independent of $\eta$, hence this gives us $\eta_0\delta_1+(1-\eta_0)\eta\sim\mathrm{DP}(\alpha\beta+\kappa\delta_1)$.

This exactly corresponds to our formulation with $\rho_1=\kappa$ and $\rho_2 = \alpha$.
\end{proof}
\section{Derivation of Gibbs Samplers}\label{supp:gibbs}
\subsection{Derivation of Direct Assignment Sampler}
\subsubsection{Step 1: Sequentially Sample $z_t, w_t, w_{t+1}$}\label{subsec:sample zw}
The joint posterior of $z_t, w_t$, and $w_{t+1}$ has the following form
\begin{equation*}
\begin{aligned}
    &\ p(z_t=k, w_t, w_{t+1} | z_{\backslash t}, w_{\backslash \{t, t+1\}}, y_{1:T}, \alpha, \beta, \{\kappa_j\}_{j=1}^{K+1}) \\
    \propto &\ p(z_t=k, w_t, w_{t+1} | z_{\backslash t}, w_{\backslash \{t, t+1\}}, \alpha, \beta, \{\kappa_j\}_{j=1}^{K+1}) \cdot p(y_t|y_{\backslash t}, z_t=k, z_{\backslash t}).\\
\end{aligned}
\end{equation*}
where $y_{\backslash t}$ are all the observations except for $y_t$, $w_{\backslash \{t, t+1\}}$ are all the $w_t$ except for $w_t,w_{t+1}$.

The predictive observation likelihood $p(y_t|y_{\backslash t}, z_t=k, z_{\backslash t})$ can be easily computed if we use a conjugate prior on the parameter in observation likelihood.  For all the emissions in the main paper, we used conjugate priors. See the normal-inverse-Wishart conjugate model in \cite{murphy2007conjugate}, Dirichlet-multinomial conjugate model in \cite{tu2014dirichlet}, Poisson-gamma conjugate model in \cite{linderman2016bayesian}, and matrix-normal inverse-Wishart conjugate model in \cite{fox2009nonparametric}.

\begin{equation}
\begin{aligned}
& p(z_t=k, w_t, w_{t+1} | z_{\backslash t}, w_{\backslash \{t, t+1\}}, \alpha, \beta, \{\kappa_j\}_{j=1}^{K+1}) \\
\propto\ &
    p(z_t=k, w_t, w_{t+1}, z_{t+1} | z_{\backslash \{t, t+1\}}, w_{\backslash \{t, t+1\}}, \alpha, \beta, \{\kappa_j\}_{j=1}^{K+1})\\
\propto\ &\int_{\bm{\pi}}p(w_t| \kappa_{z_{t-1}})p(z_t|w_t, \pi_{z_{t-1}})p(w_{t+1} | \kappa_{z_t})p(z_{t+1}|w_{t+1},\pi_{z_t}) \\
& \ \ \ \ \prod\limits_i(p(\pi_i|\alpha,\beta)\prod\limits_{\tau|z_{\tau-1}=i,w_{\tau}=0,\tau\neq t, t+1} p(z_{\tau}|\pi_i))d\bm{\pi} \\
\propto\ & \int_{\bm{\pi}}p(w_t| \kappa_{z_{t-1}})p(z_t|w_t, \pi_{z_{t-1}})p(w_{t+1} | \kappa_{z_t})p(z_{t+1}|w_{t+1},\pi_{z_t}) \\
& \ \ \ \ \prod\limits_i p(\pi_i|\{\tau|z_{\tau-1}=i,w_{\tau}=0,\tau\neq t, t+1\},\alpha,\beta)d\bm{\pi}. \label{eq:z_dist}
\end{aligned}
\end{equation}
Let $z_{t-1}=j, z_{t+1}=l$, then equation~\ref{eq:z_dist} has the following cases

\begin{equation*}\footnotesize
\begin{cases}
    \kappa_j^2 ,& \text{if }w_t = w_{t+1}=1, k=j=l \\
    (1-\kappa_j)\kappa_l\int_{\pi_j}p(z_t=l|w_t, \pi_{j}) p(\pi_j|\{\tau|z_{\tau-1}=j,w_{\tau}=0,\tau\neq t\},\alpha,\beta)d\pi_j,& \text{if }w_t =0, w_{t+1}=1,k=l \\
    (1-\kappa_j)\kappa_j\int_{\pi_j}p(z_{t+1}=l|w_{t+1}, \pi_{j}) p(\pi_j|\{\tau|z_{\tau-1}=j,w_{\tau}=0,\tau\neq t+1\},\alpha,\beta)d\pi_j,& \text{if }w_{t} =1, w_{t+1}=0, k=j \\
     (1-\kappa_j)(1-\kappa_k)\int_{\pi_j}p(z_t=k|w_t, \pi_{j}) p(\pi_j|\{\tau|z_{\tau-1}=j,w_{\tau}=0,\tau\neq t\},\alpha,\beta)d\pi_j\cdot \\
    \int_{\pi_k}p(z_{t+1}=l|w_{t+1}, \pi_{k}) p(\pi_k|\{\tau|z_{\tau-1}=k,w_{\tau}=0,\tau\neq t+1\},\alpha,\beta)d\pi_k,& \text{if }w_{t} =0, w_{t+1}=0,k\neq j \\
         (1-\kappa_j)(1-\kappa_k)\cdot\\
         \int_{\pi_j}p(z_t=j|w_t, \pi_{j})p(z_{t+1}=l|w_{t+1}, \pi_{j}) p(\pi_j|\{\tau|z_{\tau-1}=j,w_{\tau}=0,\tau\neq t,t+1\},\alpha,\beta)d\pi_j,& \text{if }w_{t} =0, w_{t+1}=0,k= j \\
    0,              & \text{otherwise } \\
\end{cases}
\end{equation*}

Similar as \cite{fox2011sticky}, we have the following equations with a close-form integration. 
\begin{align*}
    & \int_{\pi_j}p(z_t=k|w_t, \pi_{j}) p(\pi_j|\{\tau|z_{\tau-1}=j,w_{\tau}=0,\tau\neq t\},\alpha,\beta)d\pi_j 
    =  \frac{\alpha\beta_k+n_{jk}^{-t}}{\alpha+n_{j\cdot}^{-t}},\\
    & \int_{\pi_k}p(z_{t+1}=l|w_{t+1}, \pi_{k})p(\pi_k|\{\tau|z_{\tau-1}=k,w_{\tau}=0,\tau\neq t+1\},\alpha,\beta)d\pi_k
    = \frac{\alpha\beta_l+n_{kl}^{-t}}{\alpha+n_{k\cdot}^{-t}},\\
   & \int_{\pi_j}p(z_t=j|w_t, \pi_{j})  p(z_{t+1}=l|w_{t+1}, \pi_{j}) p(\pi_j|\{\tau|z_{\tau-1}=j,w_{\tau}=0,\tau\neq t,t+1\},\alpha,\beta)d\pi_j \\
    & = \frac{(\alpha\beta_j+n_{jj}^{-t})(\alpha\beta_l+n_{jl}^{-t}+\delta(j,l))}{(\alpha+n_{j\cdot}^{-t})(\alpha+n_{j\cdot}^{-t}+1)},\\
    & \int_{\pi_j}p(z_t=k|w_t, \pi_{j}) p(\pi_j|\{\tau|z_{\tau-1}=j,w_{\tau}=0,\tau\neq t\},\alpha,\beta)d\pi_j\cdot \\
    & \int_{\pi_k}p(z_{t+1}=l|w_{t+1}, \pi_{k}) p(\pi_k|\{\tau|z_{\tau-1}=k,w_{\tau}=0,\tau\neq t+1\},\alpha,\beta)d\pi_k \\
    & = \frac{\alpha\beta_k+n_{jk}^{-t}}{\alpha+n_{j\cdot}^{-t}}\frac{\alpha\beta_l+n_{kl}^{-t}}{\alpha+n_{k\cdot}^{-t}},
\end{align*}
where $n_{z_{t-1}z_t}$ is the number of transitions from state $z_{t-1}$ to state $z_t$ with $w_t=0$ in $z_{1:T}$.  $n_{z_{t-1}z_t}^{-t}$ is the number of transitions from state $z_{t-1}$ to $z_t$ with $w_t=0$, without counting the transitions $z_{t-1}$ to $z_t$ or $z_{t}$ to $z_{t+1}$. $n_{j\cdot}$ is the sum of number of transitions from $j$ to other states.

If $z_t=K+1$, i.e. a new state appears, we will increment $K$, sample self-persistence probability $\kappa_{K+1}$ for a new state from prior, and update $\beta$ in the following stick-breaking way. Sample $b\sim \mathrm{beta}(1,\gamma)$, and assign $\beta_{K}=b\beta_{k_{\mathrm{new}}}, \beta_{k_{\mathrm{new}}} = (1-b)\beta_{k_{\mathrm{new}}}$, where $\beta_{k_{\mathrm{new}}}=\sum_{i=K+1}^{\infty}\beta_i.$









\subsubsection{Step 2: Sample $\{\kappa_j\}_{j=1}^{K+1}$}\label{subsec:sample kappa}
The posterior of $\kappa_j$ is derived as follows using the beta-binomial conjugate property
\begin{equation*}
    \kappa_j \sim \mathrm{beta}(\rho_1+\sum\limits_{\tau, z_{\tau-1}=j}w_{\tau}, \rho_2+\sum\limits_{\tau, z_{\tau-1}=j}1-w_{\tau}),\ j=1,\cdots,K+1,
\end{equation*}{}
where $\kappa_{K+1}$ serves for the self-persistence probability of a new state.

\subsubsection{Step 3: Sample $\beta$}\label{subsec:sample beta}
To sample the global transition distribution $\beta$, we introduce auxiliary variables $m_{jk}$ is the auxiliary variable according to the Chinese restaurant franchise (CRF) formulation of HDP prior~\cite{teh2006}. (Note that it can be thought of as the number of tables in restaurant $j$ which serve dish $k$.) We first update $m_{jk}$, then sample $\beta$.

For each $(j, k) \in \{1,\cdots, K\}^2$, set $m_{jk} = 0, s = 0$. Then for $i = 1,\cdots, n_{jk}$, sample $x \sim \mathrm{Ber}(\frac{\alpha\beta_k}{n+\alpha\beta_k})$.  Increment $s$, and if $x = 1$ increment $m_{jk}$. Now we can sample $\beta$ as 

\begin{equation*}
    \left(\beta_1,\beta_2,\cdots,\beta_K,\beta_{k_{\mathrm{new}}}\right) \sim \mathrm{Dir}(m_{\cdot 1},\cdots,m_{\cdot K},\gamma),
\end{equation*}
where $\beta_{k_{\mathrm{new}}}=\sum_{i=K+1}^{\infty}\beta_i$ is for transiting to a new state, $m_{\cdot k}$ is $\sum_{j=1}^K m_{jk}$. A more detailed derivation of the sampling of $\beta$ can be found in \cite{fox2011sticky}.

\subsubsection{Step 4: Sample Hyperparameters $\alpha, \gamma, \rho_1, \rho_2$}\label{sec:sample hyper}

Sampling $\alpha, \gamma$ are the same as in \cite{teh2006,escobar1995bayesian}. Basically, if we apply gamma prior on $\alpha$ and $\gamma$, then by introducing some auxiliary variables, we can have the posterior of $\alpha$ and $\gamma$ to be gamma-conjugate.

For $\rho_1,\rho_2$, we follow the reparametrization trick discussed in Chapter 5 \cite{gelman2013bayesian}.  Basically, we reparametrize $\phi = \frac{\rho_1}{\rho_1+\rho_2},\eta = (\rho_1+\rho_2)^{-1/3},$ and apply a $\mathrm{Uniform}([0,1]\times[0,2])$ prior on $(\phi, \eta)$. Then we can discretize the support of $(\phi, \eta)$ and numerically compute the posterior. An alternative way to avoid the discretization would be to use Metropolis-Hasting sampling~\cite{hastings1970monte}.

\subsection{Derivation of Weak-limit Sampler}
\subsubsection{Step 1: Jointly Sample $\{z_{t},w_{t}\}_{t=1}^T$} The joint conditional distribution of $z_{1:T},w_{1:T}$ is
\begin{equation*}
    \begin{aligned}
    p(z_{1:T}, w_{1:T}| y_{1:T}, \bar{\pi}, \{\kappa_j\}_{j=1}^L, \theta) &= p(z_{T},w_{T}|z_{T-1},y_{1:T},\bar{\pi}, \{\kappa_j\}_{j=1}^L, \theta)\\
    & \ \ \ \ p(z_{T-1},w_{T-1}|z_{T-2},y_{1:T},\bar{\pi}, \{\kappa_j\}_{j=1}^L, \theta)\\
    & \cdots p(z_1|y_{1:T},\bar{\pi},\{\kappa_j\}_{j=1}^L, \theta).\\
    \end{aligned}
\end{equation*}

The conditional distribution of $z_1$ is:
\begin{equation*}
    \begin{aligned}
        p(z_1|y_{1:T},\bar{\pi},\{\kappa_j\}_{j=1}^L,\theta)\propto\  p(z_1)p(y_1|\theta_{z_1})p(y_{2:T}|z_1,\bar{\pi},\{\kappa_j\}_{j=1}^L,\theta)\\
    \end{aligned}
\end{equation*}

The conditional distribution of $z_t, t>1$ is:
\begin{equation*}
\begin{aligned}
    & p(z_t,w_t|z_{t-1}, y_{1:T},\bar{\pi},\{\kappa_j\}_{j=1}^L,\theta)\\
\propto\ & p(z_t,w_t,y_{1:T}|z_{t-1}, \bar{\pi}, \{\kappa_j\}_{j=1}^L, \theta)\\
=\ & p(z_t|\bar{\pi}_{z_{t-1}},w_t)p(w_t|z_{t-1},\{\kappa_j\}_{j=1}^L)p(y_{t:T}|z_t, \bar{\pi},\{\kappa_j\}_{j=1}^L,\theta)p(y_{1:t-1}|z_{t-1},\bar{\pi},\{\kappa_j\}_{j=1}^L,\theta)\\
\propto\ & p(z_t|\bar{\pi}_{z_{t-1}},w_t)p(w_t|z_{t-1},\{\kappa_j\}_{j=1}^L)p(y_{t:T}|z_t,\bar{\pi},\{\kappa_j\}_{j=1}^L,\theta)\\
=\ & p(z_t|\bar{\pi}_{z_{t-1}},w_t)p(w_t|z_{t-1},\{\kappa_j\}_{j=1}^L)p(y_t|\theta_{z_t})m_{t+1,t}(z_t),\\
\end{aligned}
\end{equation*}
where $m_{t+1,t}(z_t)\triangleq p(y_{t+1:T}|z_t,\bar{\pi},\{\kappa_j\}_{j=1}^L,\theta)$, which is the backward message passed from $z_t$ to $z_{t-1}$ and for an HMM is recursively defined by:

\begin{equation*}
    \begin{aligned}
        & m_{t+1,t}(z_t) = \sum_{z_{t+1},w_{t+1}} p(z_{t+1}|\bar{\pi}_{z_t},w_{t+1})p(w_{t+1}|z_t,\{\kappa_j\}_{j=1}^L)p(y_{t+1}|\theta_{z_{t+1}})m_{t+2,t+1}(z_{t+1}), t\leq T\\
        & m_{T+1,T}(z_T) = 1 \\
    \end{aligned}
\end{equation*}

\subsubsection{Step 2: Sample $\{\kappa_j\}_{j=1}^L$} Same as \ref{subsec:sample kappa} step 2, but for $j=1,\cdots,L$.

\subsubsection{Step 3: Sample $\{\beta_j\}_{j=1}^L,\{\bar{\pi}_j\}_{j=1}^L$}
\begin{equation*}
\begin{aligned}
    \beta|m,\gamma &\sim \mathrm{Dir}\left(\gamma/L+m_{\cdot1},\cdots,\gamma/L+m_{\cdot L}\right),\\
     \bar{\pi}_j | z_{1:T},w_{1:T},\alpha,\beta &\sim \mathrm{Dir}(\alpha\beta_1+n_{j1},\cdots,\alpha\beta_L+n_{jL}),\  j=1,\cdots,L.
\end{aligned}
\end{equation*}

\subsubsection{Step 4: Sample $\{\theta_j\}_{j=1}^L$} We sample $\theta$ from the posterior distribution depending on the emission function and the base measure $H$ of the parameter space $\mathrm{\Theta}$, i.e. $\theta_j | z_{1:T}, y_{1:T} \sim p(\theta_j| \{y_t|z_t = j\})$. For all the emissions in the main paper, we can easily sample $\theta_j$ from its posterior using the conjugacy properties.

\subsubsection{Step 5: Sample Hyperparameters $\alpha, \gamma, \rho_1, \rho_2$} Same as \ref{sec:sample hyper} step 4.

\section{Details of Prior in ARHMM}\label{supp:arhmm}
The MNIW prior we used in ARHMM is given by placing a matrix-normal prior $\mathcal{MN}\left(M,\Sigma_j,V\right)$ on $A_j$ given $\Sigma_j$:
\begin{equation*}
    p\left(A_j|\Sigma_j\right) = \frac{1}{(2\pi)^{d^2/2}|V|^{d/2}|\Sigma_j|^{d/2}}\exp\left(-\frac{1}{2}tr\left[\left(A_j-M\right)^{\top}\Sigma^{-1}_j\left(A_j-M\right)V^{-1}\right]\right),
\end{equation*}
where $M$ is $d\times d$ matrix, $\Sigma_j, V$ are $d\times d$ positive-definite matrix, d is the dimension of observation $y_t$; and an inverse-Wishart prior $IW(S_0,n_0)$ on $\Sigma_j$:
\begin{equation*}
    p\left(\Sigma_j\right) = \frac{|S_0|^{n_0/2}}{2^{n_0d/2}\Gamma_d(n_0/2)}|\Sigma_j|^{-(n_0+d+1)/2}\exp\left(-\frac{1}{2}tr\left(\Sigma^{-1}_j S_0\right)\right),
\end{equation*}{}
where $\Gamma_d(\cdot)$ is the multivariate gamma function. 

We set $M={\bf{0}}, V=I_{d\times d}, n_0=d+2,S_0 =0.75\bar{\Sigma},$ where $\bar{\Sigma}=\frac{1}{T}\sum_{t=1}^T(y_t-\bar{y})(y_t-\bar{y})^{\top}$ when segmenting the mouse behavior video.

\section{Simulation Results for Gaussian Emission}\label{supp:gaussian}
\begin{figure}[!ht]
    \centering
    \includegraphics[width = 1\textwidth,clip=true]{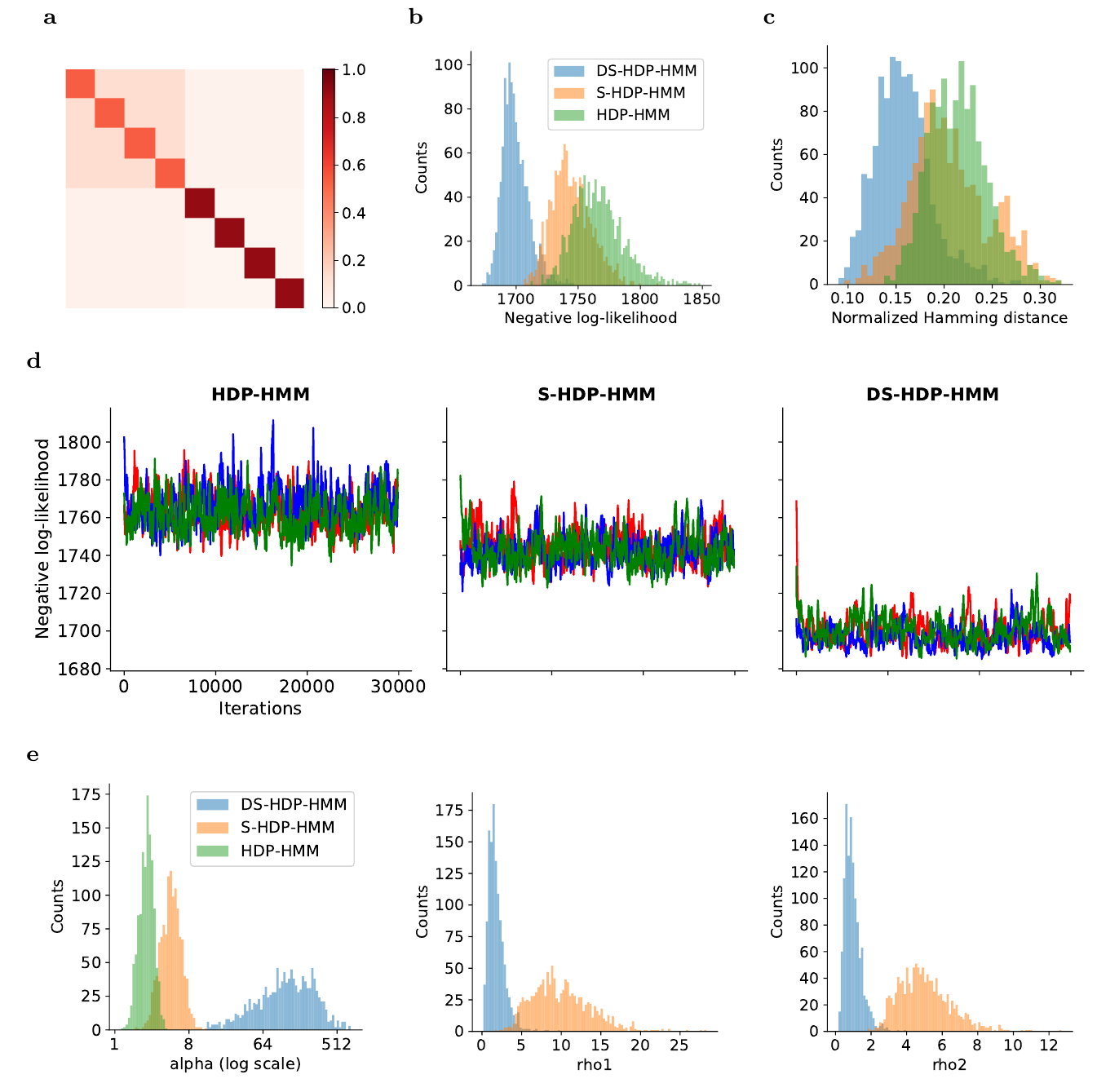}
    \caption{Results for simulated data with different self-persistence ($\rho_1,\rho_2$ small), same transition ($\alpha$ large) and Gaussian emission. We ran 3 chains, each with 30000 iterations, with 26000 iterations as burn-in for this data. Conventions as in  Figure~\ref{fig:multinomial_same_trans} in the main paper. Our model learns a big $\alpha$ and small $\rho_1,\rho_2$, which is consistent with the data.}
    \label{fig:gaussian_same_trans}
\end{figure}
We simulated data using a transition matrix (Figure~\ref{fig:gaussian_same_trans}(a)) with different $\kappa_j$ and the same $\bar{\pi}_j$ across states, which corresponds to a large transition concentration parameter $\alpha$ (big feature 1), and small self-persistence parameters $\rho_1,\rho_2$ (small feature 3).
To generate Gaussian observations, we assumed that $y_t\sim \mathcal{N}(\theta_{z_t}, 0.5^2), \theta_j\overset{iid}{\sim} \mathcal{N}(3.5, 6^2),\ j=1,\cdots,8$. We placed a Gaussian prior on the Gaussian mean parameters, with mean and variance equal to the empirical mean and variance, and fixed the observation noise as $\mathcal{N}(0,0.5^2)$. We ran 3 MCMC chains, each with 30000 iterations, with 26000 iterations as burn-in. See Figure~\ref{fig:gaussian_same_trans} for the results.

\end{document}